\documentclass{article}

\usepackage{arxiv}

\usepackage[utf8]{inputenc} % allow utf-8 input
\usepackage[T1]{fontenc}    % use 8-bit T1 fonts
\usepackage{hyperref}       % hyperlinks
\usepackage{url}            % simple URL typesetting
\usepackage{booktabs}       % professional-quality tables
\usepackage{amsfonts}       % blackboard math symbols
\usepackage{nicefrac}       % compact symbols for 1/2, etc.
\usepackage{microtype}      % microtypography
\usepackage{lipsum}
\usepackage{graphicx}
\graphicspath{ {./images/} }

\usepackage[utf8]{inputenc} % allow utf-8 input
\usepackage[T1]{fontenc}    % use 8-bit T1 fonts
\usepackage{hyperref}       % hyperlinks
\usepackage{url}            % simple URL typesetting
\usepackage{booktabs}       % professional-quality tables
\usepackage{amsfonts}       % blackboard math symbols
\usepackage{nicefrac}       % compact symbols for 1/2, etc.
\usepackage{microtype}      % microtypography
\usepackage{xcolor}         % colors

\usepackage{graphicx}
\usepackage{algorithm}
\usepackage{amsmath}
\usepackage{amssymb}
\usepackage{float}
\usepackage{pifont}
\usepackage{algpseudocode}
\usepackage{natbib}
\usepackage{hyperref}

\title{Privileged Foresight Distillation: Zero-Cost Future
Correction for World Action Models}

\author{
 Pengcheng Fang \\
  The University of Southampton\\
   \And
 Hongli Chen \\
  The University of Queensland\\
  \And
 Xiaohao Cai \\
  The University of Southampton\\
}

\begin{document}
\maketitle

\newcommand{\xmark}{\ding{55}}

% =================================================================
% FastWAM-PFD — Abstract / Introduction / Related Work
% Round 5 consolidated (after 2 adversarial reviewer passes)
% Aligned with finalized Methods (methods.tex) and Appendix A (appendix_adapter.tex)
% =================================================================

\begin{abstract}
World action models jointly predict future video and action during training, raising an open question about what role the future-prediction branch actually plays. A recent finding shows that this branch can be removed at inference with little to no loss on common manipulation benchmarks, suggesting that future information may act merely as a regularizer on the shared visual backbone. We propose instead that joint training induces an action-conditioned correction that privileged future observations impose on action denoising, and that current-only policies capture this correction only partially. Making the account precise, we formulate privileged foresight as a residual in the action-denoising direction---the difference between what a model predicts given the true future and what it predicts given only the current frame---and introduce \emph{Privileged Foresight Distillation (PFD)}, which transfers this residual from a training-time teacher into a small adapter on a current-only student. The teacher and student share the same backbone and differ only in the attention mask over video tokens; future video is never generated at inference. Controlled experiments support that this gain reflects a future-conditioned correction rather than a side effect of capacity or regularization. Empirically, PFD improves over Fast-WAM on LIBERO and RoboTwin manipulation benchmarks while preserving the current-only inference interface with only a slight adapter-induced latency overhead. This view reframes the role of future information in world action models: not as a target to predict, nor as a regularizer to absorb, but as a compressible correction to be distilled. Code is available at \href{https://github.com/PengchengFang-cs/PFD}{github.com/PengchengFang-cs/PFD}.
\end{abstract}

\section{Introduction}
\label{sec:intro}

Joint prediction of future video and action is a central design pattern in world action models, motivated by the intuition that visual foresight during training helps an agent choose better actions. A recent finding challenges this premise: a model trained jointly with video prediction can be deployed without test-time future generation while matching or exceeding the predictive variant~\citep{yuan2026fastwam}. The result has been read as evidence that test-time future imagination is unnecessary---but it leaves a deeper question unanswered. If the future branch is not used at inference, what role does future information play during training, and is any of its action-specific content lost when the branch is removed?

Two readings of this finding are possible. On a \emph{regularizer} reading, future video shapes the shared visual backbone but contributes nothing action-specific; the current-only policy captures everything useful, and there is nothing to recover. On a \emph{privileged-foresight} reading, future video induces a structured correction on the action-denoising direction itself---a correction that joint training transfers only partially to the current-only path. The two readings are observationally similar in the existing literature, yet they imply opposite methodological prescriptions: the first directs effort toward stronger visual backbones, the second toward better mechanisms for transferring the privileged signal. We find that the first reading is incomplete. Simply exposing the current-only policy to more training capacity---na\"ive finetuning of the same backbone layers---fails to improve performance (Section~\ref{sec:exp:probes}), so the gap between what joint training can teach and what the current-only policy learns is not a capacity gap. The interesting signal, if it exists, must lie in a direction that pure supervision on the action target does not reach.

We locate this signal by asking what privileged access to the future would change in the action-denoising process. During training, we instantiate the same backbone as two parallel paths, identical except for the attention mask over video tokens: a current-only student that sees only the current frame (matching the standard joint-training setup), and a privileged teacher that attends to the full future video. The teacher's action-velocity prediction minus the student's defines a \emph{foresight residual}---the component of the denoising direction that becomes predictable once future information is available. \emph{Privileged Foresight Distillation (PFD)} trains a small adapter on the student path to predict this residual from current-only context. The residual target is detached before use, so the inherited joint-training objective is not pulled away from the action target by a moving teacher signal. At inference, the teacher is discarded and the adapter augments the student's prediction at each denoising step; the current-only inference interface is preserved exactly, with the foresight-induced correction restored through a residual head whose added latency is slight (Section~\ref{sec:experiments}).

We design controlled experiments to interpret PFD's gain, isolating it from confounds of capacity, regularization, and fine-tuning-budget reallocation. None of these alternatives accounts for the observed effect, supporting a specific reading of the transferred signal: privileged foresight is a future-conditioned correction that is not recovered by matched direct fine-tuning under the same budget, and a small adapter is sufficient to absorb it.

\noindent\textbf{Contributions.} We make the following contributions.
\begin{itemize}
\item \textbf{A new perspective on future information.} We propose that future information in world action models is best understood as an action-conditioned correction residual---a direction not recovered by matched direct fine-tuning under the same budget.
\item \textbf{PFD.} We introduce a training-only teacher--student construction that makes this view operational: the teacher accesses real future during training, a small adapter distills the teacher-minus-student residual, and the adapter preserves the current-only inference interface with no future generation at test time.
\item \textbf{Controlled evidence for the transferred signal.} We design experiments that isolate PFD's gain from confounds of capacity, auxiliary regularization, and budget reallocation between backbone fine-tuning and adapter capacity, supporting the reading of the foresight residual as a future-conditioned correction.
\item \textbf{Empirical results.} PFD improves over the Fast-WAM backbone on LIBERO and RoboTwin, matching or exceeding several methods that rely on embodied pretraining, while adding only a slight inference overhead from the adapter.
\end{itemize}

\section{Related Work}
\label{sec:related}

\noindent\textbf{World action models and future video.}
Recent robot policies combine video backbones with action heads, either by jointly predicting future frames and actions~\citep{wu2024gr1, cheang2024gr2, hu2025vpp} or by conditioning actions on externally generated future videos~\citep{du2023unipi, black2024susie}. In both settings, future materialization---as pixels or latent rollouts---is required at inference and often dominates computation. Fast current-only policies remove this test-time future generation with a single forward pass~\citep{yuan2026fastwam}. We ask whether action-relevant future information can still benefit such current-only policies, and in what form.

\noindent\textbf{Uses of future information.}
Prior work mainly uses future information in two ways. \emph{Future-as-prediction} explicitly generates future frames for action conditioning~\citep{du2023unipi, black2024susie}, while \emph{future-as-representation} learns latent imagination rollouts for planning or representation learning~\citep{hafner2023dreamerv3, schrittwieser2020muzero, hansen2024tdmpc2}. Both require future content to exist at test time in some form. In contrast, PFD uses \emph{future-as-correction}: future video is available only during training, where it reveals what a current-only policy misses, and is distilled into a residual correction that is not reconstructed at inference.

\noindent\textbf{Privileged information and adapter heads.}
PFD builds on asymmetric teacher--student learning with privileged information~\citep{vapnik2009lupi, chen2019lbc}, commonly used to transfer supervision from more informed teachers to constrained students. Here, teacher and student share the same backbone parameters and differ only in their attention mask over video tokens, removing architectural confounds. Moreover, PFD defines the adapter target as the teacher--student residual rather than replacing the student with full teacher imitation; a weak teacher-consistency term is used only to stabilize the corrected output, isolating the component attributable to future access. This residual is carried by a small action-stream adapter; unlike generic parameter-efficient adapters~\citep{hu2022lora}, it is explicitly sized and trained to encode the foresight residual.

\begin{figure}[htbp]
    \centering
    \includegraphics[width=0.95\linewidth]{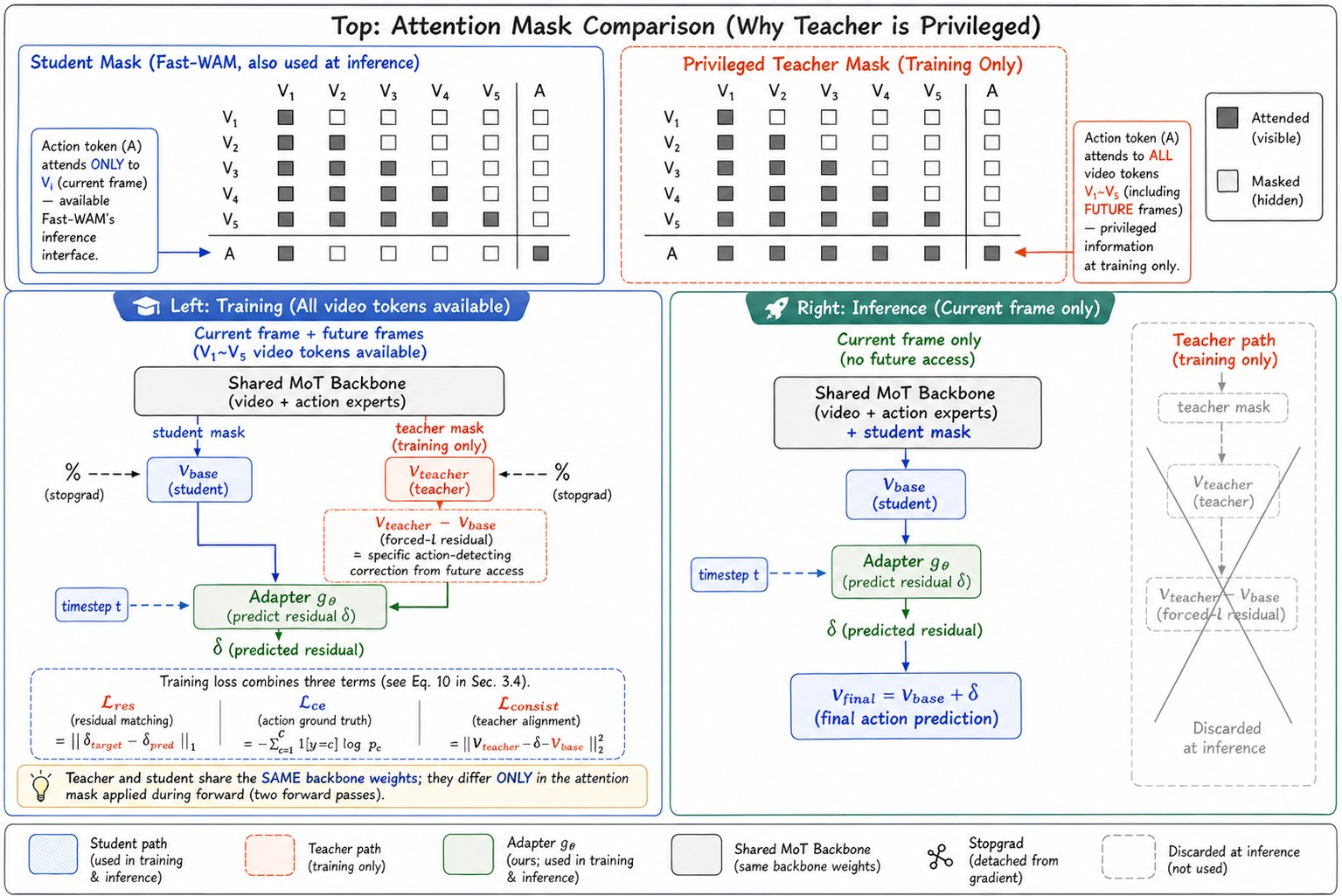}
    \caption{
    \textbf{PFD (Privileged Foresight Distillation).} \textbf{Top:} Student and privileged teacher paths differ only in their attention mask:
    the student action tokens attend to the current-frame video tokens and action tokens,
    matching the Fast-WAM current-only inference interface, whereas the teacher action
    tokens attend to all video tokens, including real future frames available only during
    training.
    \textbf{Left:} During training, the same MoT backbone is evaluated under both masks,
    yielding a live student prediction $v_{\mathrm{base}}$ and a stop-gradient privileged
    prediction $v_{\mathrm{teacher}}$. The detached residual target
    $r=\mathrm{sg}(v_{\mathrm{teacher}}-v_{\mathrm{base}})$ captures the action-denoising
    correction induced by future access. A small adapter $g_\phi$ takes the live
    $v_{\mathrm{base}}$ and predicts $\hat{\delta}$ to fit this residual, producing
    $v_{\mathrm{final}}=v_{\mathrm{base}}+\hat{\delta}$.
    \textbf{Right:} At inference, the teacher path and future video tokens are discarded.
    The model runs only the student mask and applies the adapter correction, preserving
    the current-only Fast-WAM interface with no test-time future generation and slight
    adapter-only compute cost.
    }
    \label{fig:main}
\end{figure}

\section{Method}
\label{sec:method}

PFD is a training-time mechanism that operationalizes the future-as-correction view of \S\ref{sec:intro}: a privileged path with future access produces an action-side correction signal, and a small adapter on the current-only path absorbs it. Inference uses the current-only path and the adapter; the privileged path is not instantiated.

\subsection{Preliminaries}
\label{sec:method:prelim}

Let $X = (X_1, \ldots, X_T)$ denote a sequence of $T$ video frames with $X_1$ the current frame, and let $A$ denote the corresponding action chunk. Following~\citet{yuan2026fastwam}, we adopt a Mixture-of-Transformers backbone with parameters $\theta$, comprising a video expert and an action expert, with cross-stream information exchange controlled by a joint attention mask.

Both streams are trained with flow matching. We sample timesteps $\tau_v, \tau_a \in [0,1]$ independently and draw independent Gaussian noise $\varepsilon_v, \varepsilon_a$, yielding the corrupted inputs
\begin{equation}
X_{\tau_v} \;=\; (1-\tau_v)\,\varepsilon_v + \tau_v\, X, \qquad A_{\tau_a} \;=\; (1-\tau_a)\,\varepsilon_a + \tau_a\, A,
\label{eq:corrupted-inputs}
\end{equation}
with action-velocity target $v_\mathrm{target} = A - \varepsilon_a$. We write $u_\mathrm{video}$ for the video-velocity output and $u_\mathrm{target} = X - \varepsilon_v$ for its target. We write
\[
v_\mathrm{act}\!\left(X_{\tau_v},\, A_{\tau_a},\, \tau_v,\, \tau_a;\; M\right)
\]
for the action-velocity output produced by the backbone when the joint self-attention is restricted by mask $M$. The current-only forward used at inference adopts the \emph{student mask} $M_\mathrm{S}$, under which each action-token query attends to the current-frame video tokens $X_1$ and to the other action tokens:
\begin{equation}
v_\mathrm{base} \;=\; v_\mathrm{act}\!\left(X_{\tau_v},\, A_{\tau_a},\, \tau_v,\, \tau_a;\; M_\mathrm{S}\right).
\label{eq:v-base}
\end{equation}
Throughout the displayed equations we omit first-frame observation conditioning and the per-timestep scheduler weighting on the velocity outputs for clarity; both follow~\citet{yuan2026fastwam} and are applied identically to the student and teacher forwards introduced below.

\subsection{A privileged forward via attention masking}
\label{sec:method:privileged}

PFD adds a second action forward, identical to \eqref{eq:v-base} in every respect except for the attention mask:
\begin{equation}
v_\mathrm{teacher} \;=\; \operatorname{sg}\!\left[\, v_\mathrm{act}\!\left(X_{\tau_v},\, A_{\tau_a},\, \tau_v,\, \tau_a;\; M_\mathrm{T}\right) \,\right].
\label{eq:v-teacher}
\end{equation}
The \emph{teacher mask} $M_\mathrm{T}$ allows each action-token query to attend to the full set of video tokens $X_1, \ldots, X_T$, including future frames; $\operatorname{sg}(\cdot)$ denotes the stop-gradient operator. The teacher and student forwards share the same parameters $\theta$ at every step---there is no exponential moving average, no frozen copy, and no distinct teacher network---and they consume the same noisy inputs $(X_{\tau_v}, A_{\tau_a})$ generated from a single noise sample $(\varepsilon_v, \varepsilon_a)$. The two forwards differ only in the attention mask over video tokens, which isolates the effect of future access: capacity, parameterization, optimizer state, and noise realization are held identical, so any difference between $v_\mathrm{teacher}$ and $v_\mathrm{base}$ is attributable to the enlargement of the action queries' attention support. The stop-gradient on \eqref{eq:v-teacher} further removes the teacher from the optimization graph, so it contributes no parameter update of its own and serves only as a target source for the residual we now define. Because the two forwards share $\theta$, the residual $v_\mathrm{teacher} - v_\mathrm{base}$ is a model-dependent foresight-induced correction at the current parameter state and evolves during training.

\subsection{The foresight residual and the residual adapter}
\label{sec:method:residual}

Rather than have the student imitate $v_\mathrm{teacher}$ in full, PFD distills only the component that future access changes. We define the \emph{foresight residual} as
\begin{equation}
r \;:=\; \operatorname{sg}\!\left(v_\mathrm{teacher} - v_\mathrm{base}\right);
\label{eq:residual}
\end{equation}
since $v_\mathrm{teacher}$ is already detached, this is equivalent to $r = v_\mathrm{teacher} - \operatorname{sg}(v_\mathrm{base})$. Targeting $r$ rather than $v_\mathrm{teacher}$ confines supervision to the component that future access changes at the current $\theta$; whenever the two masks induce identical predictions, the target is zero and no teacher signal enters the loss.

The residual is absorbed by a small \emph{residual adapter} $g_\varphi$ placed at the output of the action expert and applied token-wise:
\begin{equation}
\hat{\delta} \;=\; g_\varphi\!\left(v_\mathrm{base},\, \tau_a\right), \qquad v_\mathrm{final} \;=\; v_\mathrm{base} + \hat{\delta}.
\label{eq:adapter}
\end{equation}
First, the adapter consumes $v_\mathrm{base}$ rather than $\operatorname{sg}(v_\mathrm{base})$; only the residual target $r$ in \eqref{eq:residual} is detached. As we discuss in \S\ref{sec:method:loss}, this asymmetry is what allows residual supervision to influence the backbone subset $\theta'$ at all under partial fine-tuning, rather than reducing to a pure adapter-fitting problem on $\varphi$. Second, the adapter's output projection is zero-initialized, so $\hat{\delta} \equiv 0$ at the start of training and $v_\mathrm{final} = v_\mathrm{base}$ identically; the corrected forward equals the standard Fast-WAM student forward at initialization, and any departure from it accumulates only as training drives $g_\varphi$ to fit $r$.

\subsection{Training objective and gradient routing}
\label{sec:method:loss}

Let $\theta' \subseteq \theta$ denote the subset of backbone parameters that are permitted to update; the adapter parameters $\varphi$ always update. PFD trains $(\theta', \varphi)$ against the inherited video flow-matching loss together with three action-side losses:
\begin{align}
\mathcal{L}_\mathrm{video}   &= w_v(\tau_v)\,\bigl\lVert u_\mathrm{video} - u_\mathrm{target} \bigr\rVert^2, \label{eq:loss-video}\\
\mathcal{L}_\mathrm{gt}      &= w_a(\tau_a)\,\bigl\lVert v_\mathrm{final} - v_\mathrm{target} \bigr\rVert^2, \label{eq:loss-gt}\\
\mathcal{L}_\mathrm{res}     &= \bigl\lVert \hat{\delta} - r \bigr\rVert^2, \label{eq:loss-res}\\
\mathcal{L}_\mathrm{teacher} &= \bigl\lVert v_\mathrm{final} - v_\mathrm{teacher} \bigr\rVert^2, \label{eq:loss-teacher}
\end{align}
where $w_v(\cdot)$ and $w_a(\cdot)$ are the per-timestep weighting schedules of~\citet{yuan2026fastwam}. The full PFD objective is
\begin{equation}
\mathcal{L} \;=\; \lambda_\mathrm{video}\,\mathcal{L}_\mathrm{video} \;+\; \lambda_\mathrm{gt}\,\mathcal{L}_\mathrm{gt} \;+\; \lambda_\mathrm{res}\,\mathcal{L}_\mathrm{res} \;+\; \lambda_\mathrm{teacher}\,\mathcal{L}_\mathrm{teacher},
\label{eq:total}
\end{equation}
with non-negative scalar coefficients; the values used in all experiments are reported in \S\ref{sec:exp:setup}.

\noindent\textbf{Gradient routing.} The teacher forward is fully stop-gradiented at \eqref{eq:v-teacher} and contributes no update to $\theta$. The residual target $r$ in \eqref{eq:residual} is also detached, which prevents $\mathcal{L}_\mathrm{res}$ from being trivially reduced by moving its target instead of fitting it. However, because the adapter input in \eqref{eq:adapter} is the live $v_\mathrm{base}$ rather than $\operatorname{sg}(v_\mathrm{base})$, $\mathcal{L}_\mathrm{res}$ is not confined to updating $\varphi$: under partial fine-tuning, gradient also flows from $\mathcal{L}_\mathrm{res}$ through the dependence of $\hat{\delta}$ on $v_\mathrm{base}$ and into $\theta'$. Residual supervision therefore reshapes both the correction head and the backbone's emitted current-only velocity, with the detached target ensuring that the reshaping pulls $v_\mathrm{base}$ toward the privileged prediction rather than away from it.

\noindent\textbf{$\mathcal{L}_\mathrm{res}$ versus $\mathcal{L}_\mathrm{teacher}$.} The two teacher-derived losses coincide in forward value but differ in gradient path. Substituting $v_\mathrm{final} = v_\mathrm{base} + \hat{\delta}$ into \eqref{eq:loss-teacher} gives $\lVert \hat{\delta} - (v_\mathrm{teacher} - v_\mathrm{base}) \rVert^2$, which equals $\mathcal{L}_\mathrm{res}$ as a number. They diverge once gradients are computed: $\mathcal{L}_\mathrm{res}$ uses a fully detached residual target and routes gradient primarily through $g_\varphi$, with a secondary path into $\theta'$ via the adapter's dependence on $v_\mathrm{base}$; $\mathcal{L}_\mathrm{teacher}$ keeps the live $v_\mathrm{base}$ inside $v_\mathrm{final}$ on the prediction side and detaches only $v_\mathrm{teacher}$, so its gradient pulls the current-only velocity itself toward the privileged prediction rather than routing through $g_\varphi$. PFD retains both terms: $\mathcal{L}_\mathrm{res}$ supervises the adapter through a detached target, while $\mathcal{L}_\mathrm{teacher}$ pulls $v_\mathrm{base}$ toward $v_\mathrm{teacher}$ through the live prediction path.

PFD admits two regimes via the choice of $\theta'$: \emph{adapter-only} ($\theta' = \varnothing$, the backbone is frozen) and \emph{partial fine-tuning} ($\theta'$ unfreezes the last $K_a$ blocks of the action expert and the last $K_v$ blocks of the video expert); specific values and the default configuration are reported in \S\ref{sec:exp:setup}.

\subsection{Inference}
\label{sec:method:inference}

At inference, PFD preserves the current-only denoising interface of Fast-WAM. At each flow-matching denoising step, the model computes the student velocity $v_\mathrm{base}$ from \eqref{eq:v-base} under the student mask $M_\mathrm{S}$, applies the residual adapter, and uses
\[
v_\mathrm{final} \;=\; v_\mathrm{base} + g_\varphi(v_\mathrm{base},\, \tau_a)
\]
for the sampling update on $A_{\tau_a}$. The teacher mask $M_\mathrm{T}$ is never instantiated at inference, and the future video frames $X_2, \ldots, X_T$ are neither generated nor consumed. The only added cost relative to Fast-WAM is one forward pass through $g_\varphi$ per denoising step.

\section{Experiments}
\label{sec:experiments}

\subsection{Experimental setup}
\label{sec:exp:setup}
\noindent\textbf{Benchmarks.}
We evaluate on LIBERO~\citep{liu2023libero} and RoboTwin~2.0~\citep{mu2025robotwin}, following Fast-WAM~\citep{yuan2026fastwam}. LIBERO contains four suites (Spatial, Object, Goal, Long); for each, we train one model on $500$ demonstrations over $10$ tasks and report success rate over $500$ trials. RoboTwin~2.0 is a bimanual dual-arm benchmark; we use its multi-task setup with $2{,}500$ clean-scene and $25{,}000$ randomized-scene demonstrations across more than $50$ tasks, reporting success over $100$ trials per task in each condition.

\noindent\textbf{Baselines.}
Our primary baseline is Fast-WAM in two forms: ``Fast-WAM (released)'' directly transcribes the numbers from~\citet{yuan2026fastwam}, while ``Fast-WAM (reproduced)'' is re-trained with the released configuration, codebase, and schedule used by our PFD runs, and serves as the reference for reported gains. The reproduced numbers are slightly lower than the original report but follow consistent suite-level trends under a unified evaluation pipeline. For broader context, we also include published numbers for OpenVLA~\citep{kim2024openvla}, $\pi_0$~\citep{black2024pi0}, $\pi_{0.5}$~\citep{intelligence2025pi05}, Motus~\citep{motus2025}, and LingBot-VA~\citep{lingbotva2025}, taken verbatim from~\citet{yuan2026fastwam}. These five context baselines use embodied pretraining (``Emb.\ PT.''), whereas Fast-WAM and PFD use the Wan2.2-5B backbone without embodied pretraining.

\noindent\textbf{Training.}
We train for $30$ epochs on LIBERO and $15$ on RoboTwin using $8$ H100 GPUs, matching Fast-WAM's batch size, schedule, and optimizer family. We use AdamW with cosine decay, weight decay $0.01$, gradient clipping $1.0$, and benchmark-specific learning rates following Fast-WAM defaults: $6{\times}10^{-5}$ for LIBERO and $1{\times}10^{-4}$ for RoboTwin. PFD adds only the privileged forward of \S\ref{sec:method:privileged}, which shares backbone parameters and introduces one additional attention pass per step.

\noindent\textbf{Inference.}
Following Fast-WAM, we use $10$ flow-matching denoising steps with classifier-free guidance scale $1.0$. At each step, PFD runs one student forward under $M_\mathrm{S}$ and applies the residual adapter,
$v_\mathrm{final}=v_\mathrm{base}+g_\varphi(v_\mathrm{base},\tau_a)$,
as in \S\ref{sec:method:inference}. The teacher mask $M_\mathrm{T}$ is never instantiated at inference, and no future video frames are generated or consumed. End-to-end latency is reported in \S\ref{sec:exp:latency}.

\noindent\textbf{Implementation.}
All main results use the partial fine-tuning regime of \S\ref{sec:method:loss}, with trainable parameters $\theta'\cup\varphi$. Here, $\theta'$ contains the last $K_a$ action-expert blocks and last $K_v$ video-expert blocks, each expert having $30$ blocks, and $\varphi$ denotes the adapter. We set $(K_a,K_v)=(12,12)$ for both benchmarks, unfreezing about $40\%$ of blocks per expert. The adapter $g_\varphi$ is a three-layer SiLU MLP of width $512$; it takes a linear projection of the live base-action velocity $v_\mathrm{base}$ from \eqref{eq:v-base} and a sinusoidal embedding of $\tau_a$ broadcast over tokens, with zero-initialized output projection. Loss weights are fixed for all PFD runs: $\lambda_\mathrm{video}=\lambda_\mathrm{gt}=1.0$, $\lambda_\mathrm{res}=0.5$, and $\lambda_\mathrm{teacher}=0.1$.

\subsection{Main results}
\label{sec:exp:main}

\noindent\textbf{LIBERO.} Table~\ref{tab:libero} reports per-suite success rates. PFD raises the LIBERO average from $96.95$ for the reproduced Fast-WAM to $98.10$, a gain of $+1.15$ on the four-suite mean. The per-suite breakdown is $+1.6$ on Spatial, $-0.2$ on Object, $+2.6$ on Goal, and $+0.6$ on Long. PFD improves on three of the four suites; on Object, where both methods exceed $99\%$, the difference of $0.2$ points is at the binomial standard-error scale of $500$-trial evaluation. The gains are most pronounced on Goal, while Long also improves over the reproduced Fast-WAM baseline. Comparing against methods that use embodied pretraining, PFD surpasses Motus ($97.7$), $\pi_{0.5}$ ($96.9$), and $\pi_0$ ($94.1$), and trails LingBot-VA ($98.5$) by $0.40$---without invoking a separate embodied pretraining stage. Adapter-only PFD ($\theta' = \varnothing$) reaches $96.60$, competitive on Spatial, Object, and Goal but below the Fast-WAM baseline on Long; we therefore adopt partial fine-tuning as the default configuration and revisit the adapter-only regime as an ablation in \S\ref{sec:exp:probes}.

\begin{table}[t]
\centering
\small
\caption{LIBERO success rate (\%) over $500$ trials per suite. ``Emb.\ PT.'' indicates embodied pretraining; ``(reproduced)'' is re-trained under our codebase. Bold marks the rows and per-suite numbers where PFD (partial fine-tune) exceeds the reproduced Fast-WAM.}
\label{tab:libero}
\begin{tabular}{lcccccc}
\toprule
\textbf{Method} & \textbf{Emb.\ PT.} & \textbf{Spatial} & \textbf{Object} & \textbf{Goal} & \textbf{Long} & \textbf{Average} \\
\midrule
OpenVLA~\citep{kim2024openvla}            & \checkmark & 84.7 & 88.4  & 79.2 & 53.7 & 76.5 \\
$\pi_0$~\citep{black2024pi0}              & \checkmark & 96.8 & 98.8  & 95.8 & 85.2 & 94.1 \\
$\pi_{0.5}$~\citep{intelligence2025pi05}  & \checkmark & 98.8 & 98.2  & 98.0 & 92.4 & 96.9 \\
Motus~\citep{motus2025}                   & \checkmark & 96.8 & 99.8  & 96.6 & 97.6 & 97.7 \\
LingBot-VA~\citep{lingbotva2025}          & \checkmark & 98.5 & 99.6  & 97.2 & 98.5 & 98.5 \\
\midrule
Fast-WAM (released)~\citep{yuan2026fastwam} & -- & 98.2 & 100.0 & 97.0 & 95.2 & 97.60 \\
Fast-WAM (reproduced)                       & -- & 97.0 & 99.4  & 96.6 & 94.8 & $96.95^{\pm 0.08}$ \\
\midrule
\textbf{PFD (partial fine-tune, ours)}      & -- & \textbf{98.6} & 99.2 & \textbf{99.2} & \textbf{95.4} & \textbf{$98.10^{\pm 0.06}$} \\
\quad PFD (adapter-only, $\theta' = \varnothing$) & -- & 97.2 & 98.8 & 96.6 & 93.8 & 96.60 \\
\bottomrule
\end{tabular}
\end{table}

\noindent\textbf{RoboTwin~2.0.} Table~\ref{tab:robotwin} reports clean-scene, randomized-scene, and average success rates. PFD reaches $93.11 / 92.69$ on clean and randomized respectively, with an average of $92.9$, improving over the Fast-WAM row by $+1.23$ on clean, $+0.91$ on randomized, and $+1.10$ on the average. PFD's $92.9$ also exceeds the strongest embodied-pretrain baseline (LingBot-VA at $92.2$) by $0.7$ despite using no embodied pretraining, and is the highest among all Wan2.2-based entries.

\begin{table}[t]
\centering
\small
\caption{RoboTwin~2.0 success rate (\%) over $100$ trials per task. ``Emb.\ PT.'' indicates embodied pretraining; ``from Wan2.2'' is re-trained on our backbone without embodied pretraining.}
\label{tab:robotwin}
\begin{tabular}{lcccc}
\toprule
\textbf{Method} & \textbf{Emb.\ PT.} & \textbf{Clean} & \textbf{Randomized} & \textbf{Average} \\
\midrule
$\pi_0$~\citep{black2024pi0}             & \checkmark & 65.92 & 58.40 & 62.2 \\
$\pi_{0.5}$~\citep{intelligence2025pi05} & \checkmark & 82.74 & 76.76 & 79.8 \\
Motus~\citep{motus2025}                  & \checkmark & 88.66 & 87.02 & 87.8 \\
\quad Motus from Wan2.2                  & --         & 77.56 & 77.00 & 77.3 \\
LingBot-VA~\citep{lingbotva2025}         & \checkmark & 92.90 & 91.50 & 92.2 \\
\quad LingBot-VA from Wan2.2             & --         & 80.60 & --    & 80.6 \\
\midrule
Fast-WAM~\citep{yuan2026fastwam}         & --         & 91.88 & 91.78 & $91.8^{\pm 0.22}$ \\
\midrule
\textbf{PFD (partial fine-tune, ours)}   & --         & \textbf{93.11} & \textbf{92.69} & \textbf{$\mathbf{92.9^{\pm 0.12}}$} \\
\bottomrule
\end{tabular}
\end{table}

\subsection{Isolating the foresight signal}
\label{sec:exp:probes}

\begin{figure}[t]
\centering
% Bar chart placeholder: probe-by-probe LIBERO average vs. Fast-WAM (reproduced) and PFD (default).
\includegraphics[width=0.78\linewidth]{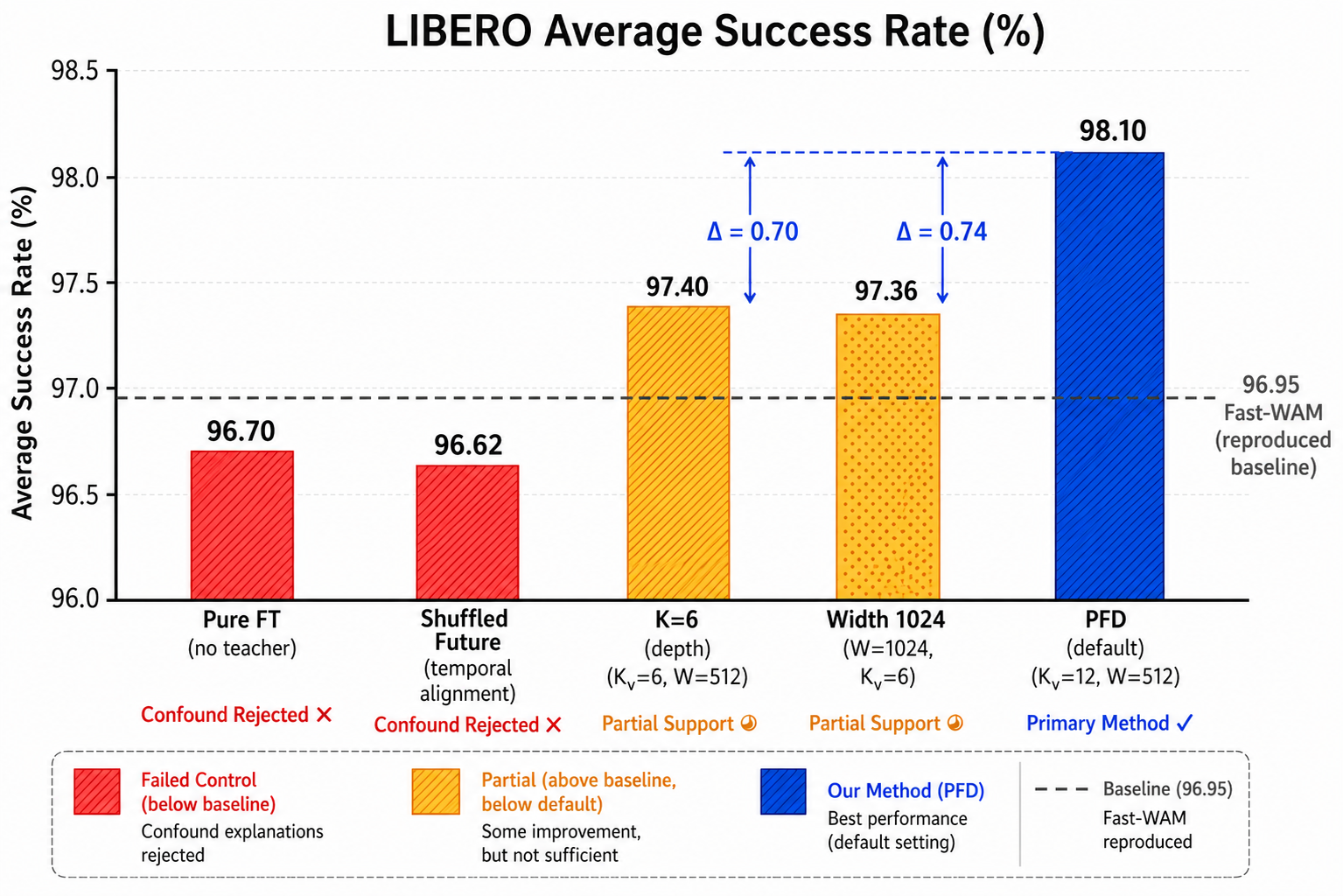}
\caption{LIBERO average success rate for the three primary probes, Fast-WAM (reproduced), and PFD (default). Breaking matched-capacity (Pure FT) or temporal-correspondence (Shuffled Future) drops accuracy below the Fast-WAM baseline (red); the budget-reallocation control---redirecting video-side fine-tuning budget to a wider adapter (Width $1024$ at $K_v = 6$)---closes only part of the PFD gap (amber). The supplementary fine-tune-depth check ($K = 6$) is shown for comparison and lands at the similar level.}
\label{fig:probes}
\end{figure}

The aggregate gain reported in \S\ref{sec:exp:main} is consistent with the privileged-foresight account but does not by itself rule out simpler explanations: extra trainable capacity in $\theta'$, generic regularization from a second teacher forward, or a different allocation of the fine-tuning budget between backbone depth and adapter capacity. We design three controlled probes that share PFD's training budget but each break exactly one ingredient of the foresight transfer, and verify whether breaking that ingredient erases the gain. We run probes on LIBERO; the four-suite split exposes capacity and correspondence dimensions independently of RoboTwin's bimanual coordination, which we treat as an end-to-end test in \S\ref{sec:exp:main}. Numerical results are collected in Table~\ref{tab:probes} and visualized in Figure~\ref{fig:probes}.

\noindent\textbf{Matched-capacity control.} The first probe, \emph{pure finetune}, unfreezes the same backbone subset $\theta' = (K_a, K_v) = (12, 12)$ and trains against the action ground truth alone, with no teacher forward and no adapter. If the PFD gain were attributable to the additional trainable capacity that $\theta'$ exposes, this control would match or exceed PFD. It does not: pure finetune scores $96.4 / 99.2 / 96.4 / 94.8$ for an average of $96.70$, which is $-0.25$ below the reproduced Fast-WAM and $-1.40$ below PFD. Unfreezing the same subset of layers under direct action supervision slightly hurts the current-only policy at this training budget. The signal that PFD transfers is therefore not accessible to direct supervision on the action target, even when the layers permitted to update are identical.

\noindent\textbf{Shuffled-future control.} The second probe, \emph{shuffled-future} PFD, replaces the teacher's future frames $X_2, \ldots, X_T$ at every training step with frames drawn from an unrelated trajectory in the same batch. The teacher mask $M_\mathrm{T}$, the adapter, the loss weights, and the schedule are otherwise identical to the default PFD run. If the gain reflected auxiliary-loss regularization or the mere presence of a second supervisory target, destroying the temporal correspondence between $X_1$ and $X_{2:T}$ should leave it largely intact, since the input statistics and loss magnitudes are preserved. Instead, shuffled-future PFD scores $96.1 / 99.2 / 96.2 / 95.0$ for an average of $96.62$, which is $-0.33$ below the reproduced Fast-WAM and $-1.48$ below PFD. The transferred signal therefore depends on genuine current-to-future correspondence, not on incidental properties of the teacher forward.

\noindent\textbf{Depth--width trade-off probe.} The third probe asks whether reducing video-side fine-tuning depth while increasing residual-head width can substitute for the default full-depth PFD configuration. We double the adapter hidden width from $512$ to $1024$ and, to test a practical depth--width trade-off, halve the video-expert fine-tune depth from $K_v = 12$ to $K_v = 6$ (the action expert is held at $K_a = 12$). This redirects the freed compute from updating the deeper video stack into a wider correction head. The resulting configuration scores $97.9 / 99.8 / 97.2 / 94.5$ for an average of $97.36$, which is $-0.74$ below the default PFD at width $512$. Redirecting fine-tuning budget from the video expert to a wider adapter therefore fails to recover the gain. We cannot rule out a clean adapter-width effect at fixed $(K_a, K_v) = (12, 12)$ and leave that question to future work.

\noindent\textbf{Reading the probes together.} Relative to the default PFD average of 98.10, the three probes land at deltas of $-1.40$ (matched-capacity), $-1.48$ (shuffled-future), and $-0.74$ (budget-reallocation). Two readings split cleanly. \emph{Ruled out}: extra trainable capacity and auxiliary-loss regularization each erase the PFD gain when isolated, with the matched-capacity row falling below the frozen baseline and the shuffled-future row matching it. \emph{Argued against, not ruled out}: budget-reallocation closes only part of the gap, and the supplementary $K = 6$ row reported below scores similarly at $97.40$ without any adapter widening, corroborating the reading that the residual gap from $97.40$ to $98.10$ tracks fine-tune depth on the video expert rather than adapter width. None of the three confounds reproduces the privileged-foresight residual $r$ in \eqref{eq:residual}: $r$ is by construction the component of the action-velocity field that becomes available when the attention mask exposes future frames at the same parameters, and the alternatives examined here change capacity, target, or budget allocation while leaving that mask alone.

\noindent\textbf{Fine-tune depth (supplementary).} As a supplementary check, we ablate the depth of the trainable backbone subset by halving $K = K_a = K_v$ from $12$ to $6$ at the default adapter width of $512$. This configuration scores $97.9 / 99.7 / 97.3 / 94.7$ for an average of $97.40$, which is $-0.70$ below default PFD but $+0.45$ above the reproduced Fast-WAM. Its close correspondence to the budget-reallocation row at 97.36 indicates that at the half-depth setting the gap to default PFD is already determined by $K$ and is not closed by adapter widening. The teacher therefore contributes useful signal even at half depth, and additional fine-tune depth on the video expert continues to absorb it.

\begin{table}[t]
\centering
\small
\caption{LIBERO epistemic probes (success rate \%). The first three probes break, in turn, capacity, temporal correspondence, and the allocation of the fine-tuning budget between backbone depth and adapter width. The last two rows are supplementary ablations on the adapter regime and on fine-tune depth.}
\label{tab:probes}
\begin{tabular}{lccccc}
\toprule
\textbf{Configuration} & \textbf{Spatial} & \textbf{Object} & \textbf{Goal} & \textbf{Long} & \textbf{Average} \\
\midrule
Fast-WAM (reproduced)                                & 97.0 & 99.4 & 96.6 & 94.8 & 96.95 \\
\midrule
Pure finetune ($\theta'$, no teacher)                & 96.4 & 99.2 & 96.4 & 94.8 & 96.70 \\
Shuffled-future PFD                                  & 96.1 & 99.2 & 96.2 & 95.0 & 96.62 \\
PFD, width $1024$ at $(K_a, K_v) = (12, 6)$          & 97.9 & 99.8 & 97.2 & 94.5 & 97.36 \\
\midrule
\textbf{PFD (default)} $(K_a, K_v) = (12, 12)$, $W = 512$ & \textbf{98.6} & 99.2 & \textbf{99.2} & \textbf{95.4} & \textbf{98.10} \\
\quad adapter-only ($\theta' = \varnothing$)         & 97.2 & 98.8 & 96.6 & 93.8 & 96.60 \\
\quad fine-tune depth $(K_a, K_v) = (6, 6)$          & 97.9 & 99.7 & 97.3 & 94.7 & 97.40 \\
\bottomrule
\end{tabular}
\end{table}

\subsection{Inference overhead}
\label{sec:exp:latency}

\begin{table}[h]
\centering
\small
\caption{
End-to-end inference latency per action chunk. In our LIBERO setting, one
chunk contains $32$ actions. Slowdown is reported relative to the current-only
Fast-WAM cached-context baseline.
}
\label{tab:overhead}
\begin{tabular}{lccc}
\toprule
\textbf{Method} & \textbf{Test-time future} & \textbf{Latency (ms/chunk)} & \textbf{Slowdown} \\
\midrule
Fast-WAM-Joint  & joint video+action denoising  & $786.2$  & $3.05\times$ \\
Fast-WAM-IDM    & generate future, then IDM      & $1098.1$ & $4.26\times$ \\
\midrule
Fast-WAM        & none (current-only)            & $257.7$  & $1.00\times$ \\
\textbf{PFD (ours)} & none (current-only) + adapter & $\mathbf{271.0}$ & $\mathbf{1.05\times}$ \\
\bottomrule
\end{tabular}
\end{table}

A central practical question for any future-aware policy is how much foresight
costs at deployment time. Models that materialize future video at inference---
either by jointly denoising future frames and actions, or by generating a future
clip before an inverse-dynamics module (IDM)---pay a multiplicative latency
penalty, because future frames must be produced before any action chunk can be
emitted. To put PFD's cost in context, we project the relative slowdowns reported
by Fast-WAM onto our measured cached-context baseline: joint denoising and
imagine-then-execute IDM inference would be approximately $3.05\times$ and
$4.26\times$ slower than the current-only interface.

Table~\ref{tab:overhead} reports end-to-end latency per action chunk on a single
H100, measured over $20$ trials after discarding $5$ warmup runs, under the
$10$ flow-matching denoising steps used throughout this paper. In our LIBERO
setup, each \texttt{infer\_action} call predicts a chunk of $32$ actions. The
current-only Fast-WAM cached-context baseline runs at $257.77$\,ms per chunk,
while PFD runs at $271.04$\,ms per chunk, adding $13.27$\,ms or $5.15\%$
overhead. We also evaluate the prompt-mode implementation of PFD, which runs at
$282.29$\,ms per chunk and adds $7.80$\,ms, or $2.84\%$, over its corresponding
prompt-mode baseline.

These results show that the adapter cost remains small under both inference
implementations. The added latency is attributable to one lightweight
$g_\varphi$ forward at each denoising step, which is slight compared with the
Wan2.2-5B backbone. PFD never instantiates the teacher mask at inference and
never generates future video frames, exactly as specified in
\S\ref{sec:method:inference}; the deployment profile of the current-only
interface is preserved while recovering the foresight-induced correction.

% =================================================================
% FastWAM-PFD — Conclusion + Limitations, v3
% Round 3 revision (per author): tighter Conclusion + 2-item Limitations.
% =================================================================

\section{Conclusion}
\label{sec:conclusion}

We revisited the role of future video in world action models once test-time imagination is removed, and argued that future is best understood not as a prediction target nor as a regularizer to absorb, but as a compressible correction to be distilled. PFD operationalizes this view with a same-backbone teacher--student construction and a small output-side adapter that absorbs the foresight residual. Three epistemic probes---matched-capacity, shuffled-future, and budget-reallocation---attribute the gain to the foresight signal itself, and PFD improves over Fast-WAM on both LIBERO and RoboTwin with a slight adapter-only inference overhead while preserving the current-only inference interface exactly.

\noindent\textbf{Limitations.} Two limitations are worth noting. First, the construction is deliberately simple: a single output-side MLP adapter, a full-horizon teacher mask, and a single backbone family. More expressive adapter designs---multi-scale, gated, or cross-attentive---and richer teacher-mask schedules remain to be explored. Second, our claims are empirical: we observe that the foresight residual is absorbable by a small adapter and that capacity, regularization, and budget-reallocation alternatives do not account for the gain, but we do not provide a formal characterization of when the residual admits a low-capacity approximation. A theoretical account would tell us a priori which task families and backbones PFD should help.

\bibliographystyle{plainnat}
\bibliography{references}

\end{document}